\documentclass[10pt,twocolumn]{article}

\AtBeginDocument{%
  }

\usepackage{times}
\usepackage{balance}
\usepackage{latexsym}
\usepackage{booktabs}
\usepackage{colortbl}
\usepackage{xcolor}
\usepackage{multirow}
\usepackage{geometry}
\usepackage{amsmath}
\usepackage{bbm}
\usepackage{bm}
\usepackage{color}
\usepackage{multirow}
\usepackage{array}
\usepackage{booktabs}
\usepackage{diagbox}
\usepackage{bbding}
\usepackage{colortbl}
\usepackage{tabularx}
\usepackage{makecell}
\usepackage{hyperref}
\usepackage{subfigure}

\usepackage{amsmath, amssymb, overpic, textpos}
\usepackage{algorithmicx}
\usepackage{algpseudocode}
\usepackage{listings}
\usepackage{multicol}
\usepackage{xspace}
\usepackage{wrapfig}
\usepackage{graphicx}
\usepackage{graphbox}
\usepackage{arydshln}
\usepackage{algorithm}
\usepackage{adjustbox}
\usepackage{kotex}
\usepackage{caption}
\usepackage{setspace} 
\usepackage{epigraph}
\usepackage{tcolorbox}
\usepackage[numbers,sort&compress]{natbib}
\usepackage{tcolorbox}
\newcommand{\ds}{{RDVSv2}\xspace}
\newcommand{\model}{{CPSAM}\xspace}
\definecolor{mygray}{gray}{.8}

\title{RDVSv2: A Large-scale Benchmark for RGB-D Video Salient Object Detection}
\author{
Tianyu Li\textsuperscript{1},
Jiahao He\textsuperscript{2},
Keren Fu\textsuperscript{1,2,*},
and Qijun Zhao\textsuperscript{1,2}\\[3pt]
\small \textsuperscript{1}National Key Laboratory of Fundamental Science on Synthetic Vision, Sichuan University, Chengdu, China\\
\small \textsuperscript{2}College of Computer Science, Sichuan University, Chengdu, China\\
\small \textsuperscript{*}Corresponding author
}
\date{}

\begin{document}
\maketitle

\begin{abstract}
We introduce \ds, a large-scale benchmark for RGB-D video salient object detection (RGB-D VSOD) with dense frame-level annotations. Existing datasets in this emerging field are often limited in scale and annotation quality, while also relying on less geometry-consistent depth cues. To address these limitations, \ds is built from publicly accessible stereoscopic online videos and contains 249 video sequences with 29,077 annotated frames. It includes depth maps derived from stereoscopic videos, together with frame-wise salient object masks annotated with eye-tracking guidance. Compared with existing datasets, \ds is much larger in scale and covers more diverse and challenging scenarios.
In addition, we establish a strong baseline for RGB-D VSOD based on Segment Anything Model 2 (SAM2). Specifically, we employ a parameter-efficient fine-tuning (PEFT) strategy to adapt the SAM2 encoder to jointly encode RGB, depth, and optical flow cues. Extensive experiments show that \ds is substantially more challenging for existing RGB-D VSOD methods. Meanwhile, the proposed baseline achieves state-of-the-art results on \ds and existing RGB-D VSOD benchmarks. We hope that \ds and the provided baseline will serve as useful resources for future research on RGB-D VSOD and related multi-modal video understanding tasks. Our dataset and code will be available at \url{https://github.com/ltynick/RDVSv2}.
\end{abstract}


\section{Introduction}
Salient object detection (SOD)~\cite{wang2021salient} aims to identify the most visually distinctive objects in a scene and has become a fundamental problem in computer vision. While substantial progress has been made in image-based SOD, the dynamic nature of videos introduces the additional challenge of temporal consistency, giving rise to video salient object detection (VSOD)~\cite{fan2019shifting}. Incorporating depth information, which provides complementary geometric and spatial cues, further extends VSOD to RGB-D VSOD~\cite{mou2024salient} and offers the potential for more robust segmentation in challenging scenarios such as occlusion and motion blur.

Despite its potential, the progress of RGB-D VSOD is hindered by the lack of large-scale, high-quality benchmarks. Current benchmarks, including DViSal~\cite{li2023dvsod}, RDVS~\cite{mou2024salient}, and ViDSOD-100~\cite{lin2024vidsod}, are limited in scale, annotation density, annotation protocols (lacking eye-tracking as reference), and scenario diversity, which collectively constrain the training and evaluation of advanced models. Meanwhile, powerful vision foundation models, particularly the Segment Anything Model (SAM) series \cite{kirillov2023segment, ravi2024sam}, have substantially advanced visual segmentation. However, their reliance on manual prompts limits their direct applicability to automatic VSOD. Their adaptation to multi-modal video inputs, such as RGB, depth, and optical flow, remains largely underexplored.

To address these challenges, we present a new benchmark dataset for RGB-D VSOD together with a baseline method. First, we introduce the \textbf{\ds} dataset, which is much larger than the earlier RDVS dataset, draws from different data sources, and has no overlap with RDVS. Built from diverse stereoscopic videos, \textbf{\ds} provides stereoscopic depth, large scale, high resolution, and dense per-frame annotations guided by eye-tracking, thereby better reflecting realistic visual attention shifts. Second, we propose a baseline (\model) that effectively adapts SAM2 \cite{ravi2024sam} for RGB-D VSOD. We design a hierarchical parameter-efficient fine-tuning (PEFT) strategy in which parallel Low-Rank Adaptation (LoRA) \cite{hu2022lora} modules capture modality-specific cues from depth and optical flow, while a shared \textbf{C}ross-modal \textbf{P}rompting adapter learns shared representations across modalities. This strategy thus enables effective tri-modal fusion within \textbf{SAM}2's frozen encoder. Extensive experiments show that \textbf{\ds} presents a substantially more challenging benchmark for existing RGB-D VSOD methods, underscoring its value for future evaluation. Meanwhile, the proposed \model achieves state-of-the-art (SOTA) performance across multiple RGB-D VSOD benchmarks, providing a strong baseline for future research.

Our main contributions are summarized as follows:
\begin{itemize}
    \item We introduce \textbf{\ds}, a large-scale benchmark for RGB-D VSOD built from stereoscopic videos, with stereo-derived depth maps and frame-wise annotations guided by eye-tracking data. Compared with existing benchmarks, \textbf{\ds} offers larger scale, higher resolution, and more diverse and challenging scenarios. The \textbf{\ds} dataset together with the \emph{supplementary material} are available at \url{https://github.com/ltynick/RDVSv2}.
    \item We establish a strong baseline, dubbed \model, by adapting SAM2 to the RGB-D VSOD task using a PEFT strategy.
    \item We establish a comprehensive benchmark on \textbf{\ds} by evaluating 11 representative models, demonstrating the value of \textbf{\ds} as a challenging benchmark and the strong performance of \model.
\end{itemize}

\section{Related Work}

\subsection{Salient Object Detection}

\textbf{RGB-D SOD and VSOD.}
RGB-D SOD leverages depth as complementary geometric cues for saliency detection, and existing methods mainly focus on cross-modal fusion and representation learning~\cite{fu2020jl, ji2021calibrated, hu2024cross, sun2023catnet}. 
VSOD extends saliency detection in dynamic scenes by modeling temporal dependencies, with existing methods often built upon optical flow, memory mechanisms, or query-based designs~\cite{li2018flow, li2019motion, oh2019video, cheng2022xmem, wang2023look, fang2024learning}.

\textbf{RGB-D VSOD and Related Datasets.}
RGB-D VSOD combines depth cues with temporal modeling, bridging RGB-D SOD and VSOD. Because RGB-D VSOD datasets were previously scarce, early studies often relied on synthetic depth maps~\cite{lu2022depth}. Recently, several benchmark datasets for RGB-D VSOD have been introduced. DViSal~\cite{li2023dvsod} provides 237 annotated RGB-D videos together with a baseline model that combines RGB-D feature fusion and temporal memory modeling. RDVS~\cite{mou2024salient} further introduces frame-wise annotations across diverse scenes and introduces DCTNet+, which improves performance through enhanced cross-modal interaction. ViDSOD-100~\cite{lin2024vidsod} provides 100 RGB-D videos together with ATF-Net, which adopts dual-stage feature aggregation for finer-grained multi-modal interaction. More recently, unified frameworks such as Samba~\cite{he2025samba} and lightweight architectures such as MFENet~\cite{suolang2025lightweight} have also been explored for RGB-D VSOD.

\subsection{Segment Anything Model}

The Segment Anything Model (SAM) series~\cite{kirillov2023segment, ravi2024sam} has demonstrated strong generalization in segmentation tasks. However, SAM heavily depends on manual prompts, which limits its direct applicability to automatic tasks such as saliency detection. Existing approaches mainly adapt SAM by generating pseudo-prompts or applying PEFT.
Several works have explored SAM-based architectures for saliency detection. MDSAM~\cite{gao2024multi} introduces multi-scale adapters to enhance SAM representations, while Sammese~\cite{wang2024adapting} integrates multi-modal information within SAM's encoder. SAMSOD~\cite{liu2026samsod} employs unimodal supervision to improve multi-modal learning, and KAN-SAM~\cite{li2025kan} incorporates thermal features via KAN adapters. Nevertheless, most of these methods focus on image-based SOD and are not designed for RGB-D VSOD. Recently, Lin et al.~\cite{lin2025sam} adapted SAM2 to RGB-D VSOD through depth-guided adaptive queries (SAM-DAQ). In contrast to SAM-DAQ, our approach incorporates optical flow as an auxiliary modality instead of relying on memory-related modules, resulting in a simpler and more general baseline.

\section{Proposed \ds Dataset}

\subsection{Dataset Construction}

\begin{figure*}[t]
\centering
\includegraphics[width=0.8\linewidth]{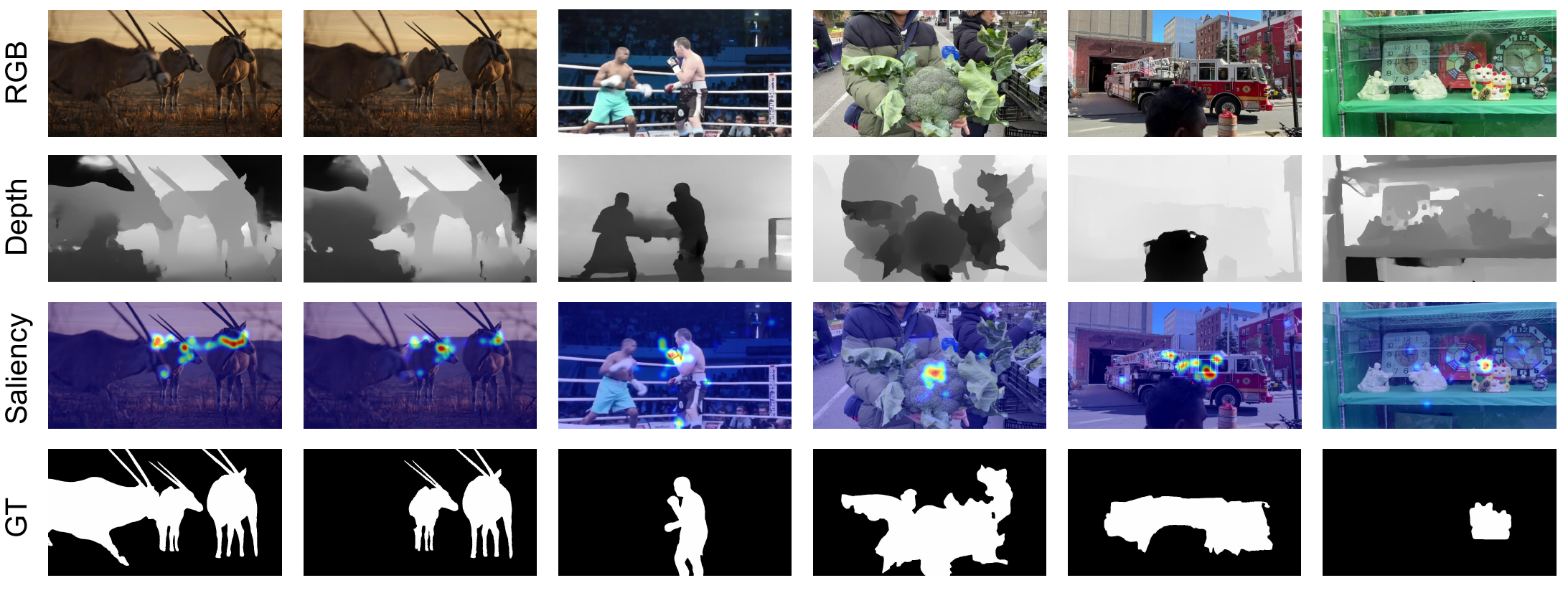}
\vspace{-0.1em}
\caption{
Illustrative frames from \ds. The two leftmost columns of frames illustrate the attention shift.
}
\label{fig:dataset}
\end{figure*}

\textbf{Data Collection.} Unlike existing RGB-D VSOD datasets~\cite{li2023dvsod, mou2024salient, lin2024vidsod}, which are typically derived from RGB-D video benchmarks, \ds is constructed from publicly accessible stereoscopic videos. This strategy enables broader scene coverage and more complex visual conditions. In total, the dataset contains 249 video sequences with 29,077 annotated frames. Since the source videos are stereoscopic, depth maps are derived from the disparity between the left and right views rather than obtained through monocular depth estimation. For each stereo pair, RAFT~\cite{teed2020raft} is used to estimate the dense correspondence between the left and right views. The horizontal flow component is then taken as a proxy for stereo disparity. After min-max normalization, the proxy map is converted into a grayscale depth image. This provides depth cues with stronger geometric consistency for RGB-D saliency analysis.

\textbf{Dataset Annotation.} We adopt the eye-tracking-based annotation protocol introduced in RDVS~\cite{mou2024salient}. Gaze data are collected from 22 participants (aged 22--35, with normal or corrected-to-normal vision) using a Tobii Eye Tracker 4C (90 Hz) during free-viewing of the video stimuli. The videos are displayed on a 23.8-inch screen (1920$\times$1080) at a viewing distance of 70 cm, with a chinrest used to minimize head movement. Only RGB frames are displayed (depth information is withheld to avoid influencing visual attention). The aggregated fixation points are converted into saliency heatmaps to guide the identification of salient objects in each frame. Additional details are provided in the \emph{supplementary material}. To improve annotation efficiency, we adopt a semi-automatic pipeline. Annotators first identify candidate salient objects according to the fixation heatmaps, after which SAM2~\cite{ravi2024sam} is used to generate initial video object segmentation masks. The resulting masks are then manually refined to obtain accurate object boundaries. This procedure naturally captures dynamic saliency shifts, where the salient object may change over time, which is consistent with DAVSOD~\cite{fan2019shifting}. Fig.~\ref{fig:dataset} presents a brief overview of the \ds dataset, including RGB frames, depth maps, saliency heatmaps, and the corresponding ground truth (GT).

\subsection{Dataset Statistics and Comparisons}
\label{sec:DSC}

As summarized in Table~\ref{tab:datasets_cmp}, we compare \ds with several commonly used video based SOD datasets. 
Our dataset offers the following advantages:

\textbf{+}\ds is substantially larger in scale than existing RGB-D VSOD datasets, which typically contain fewer than 10,000 annotated frames. In contrast, \ds contains 249 video sequences with 29,077 annotated frames, making it the largest among the RGB-D VSOD benchmarks considered here and even larger than the VSOD dataset DAVSOD~\cite{fan2019shifting}.

\textbf{+}\ds also adopts an eye-tracking-guided annotation protocol. It follows the protocol introduced in RDVS, where salient objects are determined according to fixation data collected from human observers. This differs from datasets such as DViSal and ViDSOD-100, which do not incorporate eye-tracking guidance.

\textbf{+}Despite its large scale, \ds provides dense frame-wise annotations for all video sequences. This supports temporal consistency and contrasts with some large-scale datasets such as DViSal, which do not annotate every frame.

\textbf{+}\ds also provides higher spatial resolution than previous RGB-D VSOD datasets, providing richer spatial details and making the dataset better suited for evaluating modern segmentation models.

\textbf{+}In addition, \ds provides depth maps derived from stereoscopic videos and supports attention-shift scenarios where salient objects may change over time. These characteristics are absent from several earlier VSOD datasets.

\begin{table*}[!t]
\centering
\small
\renewcommand{\arraystretch}{1.15}

\caption{\textbf{Statistical analysis of existing benchmark video-based SOD datasets and our \ds dataset. Horizontal lines separate different dataset categories: VSOD and RGB-D VSOD.}
\textbf{\#Vi.:} number of videos.
\textbf{\#AF.:} number of salient objects in the images/video frames.
\textbf{DL:} whether dense (per-frame) annotations are provided.
\textbf{AS:} whether attention shift is considered.
\textbf{EF:} whether salient objects are annotated according to eye-fixation records.
\textbf{SD:} whether stereo-derived depth is provided together with RGB frames.}
\label{tab:datasets_cmp}

\vspace{-4pt}

\begin{tabular*}{0.8\textwidth}{
    @{\extracolsep{\fill}}
    l c c c c c c c
    @{}
}
\toprule
\textbf{Dataset}
& \textbf{\#Vi.}
& \textbf{\#AF.}
& \textbf{DL}
& \textbf{AS}
& \textbf{EF}
& \textbf{SD}
& \textbf{Resolution (H$\times$W)}
\\
\midrule

SegTrack-V2 \cite{li2013video}
& 14
& 1065
& $\checkmark$
& 
& 
& 
& [212,360]$\times$[259,640]
\\

FBMS \cite{ochs2013segmentation}
& 59
& 720
& 
& 
& 
& 
& [288,540]$\times$[350,960]
\\

ViSal \cite{wang2015consistent}
& 17
& 193
& 
& 
& 
& 
& [240,288]$\times$[320,512]
\\

DAVIS \cite{perazzi2016benchmark}
& 50
& 3455
& $\checkmark$
& 
& 
& 
& [900,1080]$\times$[1600,1920]
\\

VOS \cite{li2017benchmark}
& 200
& 7467
& 
& 
& $\checkmark$
& 
& [321,800]$\times$[408,800]
\\

DAVSOD \cite{fan2019shifting}
& 226
& 23938
& $\checkmark$
& $\checkmark$
& $\checkmark$
& 
& 360$\times$640
\\

\midrule

DViSal \cite{li2023dvsod}
& 237
& 7117
& 
& $\checkmark$
& 
& $\checkmark$
& [320,1080]$\times$[640,1920]
\\

ViDSOD-100 \cite{lin2024vidsod}
& 100
& 9362
& $\checkmark$
& $\checkmark$
& 
& $\checkmark$
& 480$\times$640
\\

RDVS \cite{mou2024salient}
& 57
& 4087
& $\checkmark$
& $\checkmark$
& $\checkmark$
& $\checkmark$
& [480,720]$\times$[640,1280]
\\

\textbf{\ds (Ours)}
& \textbf{249}
& \textbf{29077}
& \textbf{$\checkmark$}
& \textbf{$\checkmark$}
& \textbf{$\checkmark$}
& \textbf{$\checkmark$}
& \textbf{[540,1080]$\times$[960,1920]}
\\

\bottomrule
\end{tabular*}

\vspace{-5pt}
\end{table*}

\begin{figure}[t]
\centering
\begin{minipage}[t]{0.59\linewidth}
    \centering
    \includegraphics[width=\linewidth]{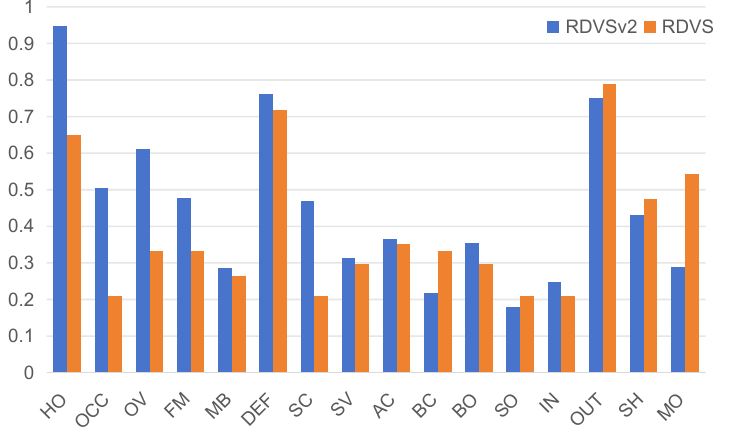}
    \vspace{-1em}
\end{minipage}
\hfill
\begin{minipage}[t]{0.39\linewidth}
    \centering
    \includegraphics[width=\linewidth]{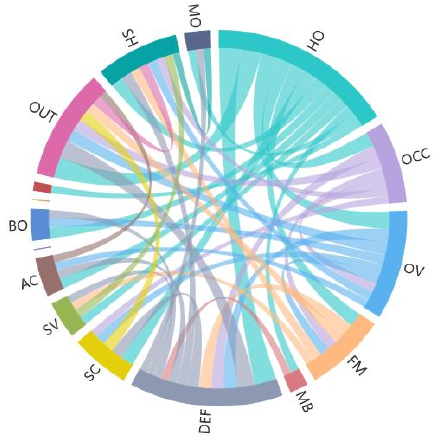}
    \vspace{-1em}
\end{minipage}
\caption{Attribute-based analyses of \ds with comparison to RDVS (left) (the counts of attributes are normalized by the total number of videos), and the pairwise dependencies across different attributes (right).}
\label{fig:combined}
\end{figure}

\begin{figure}[t]
\centering
\includegraphics[width=0.99\linewidth]{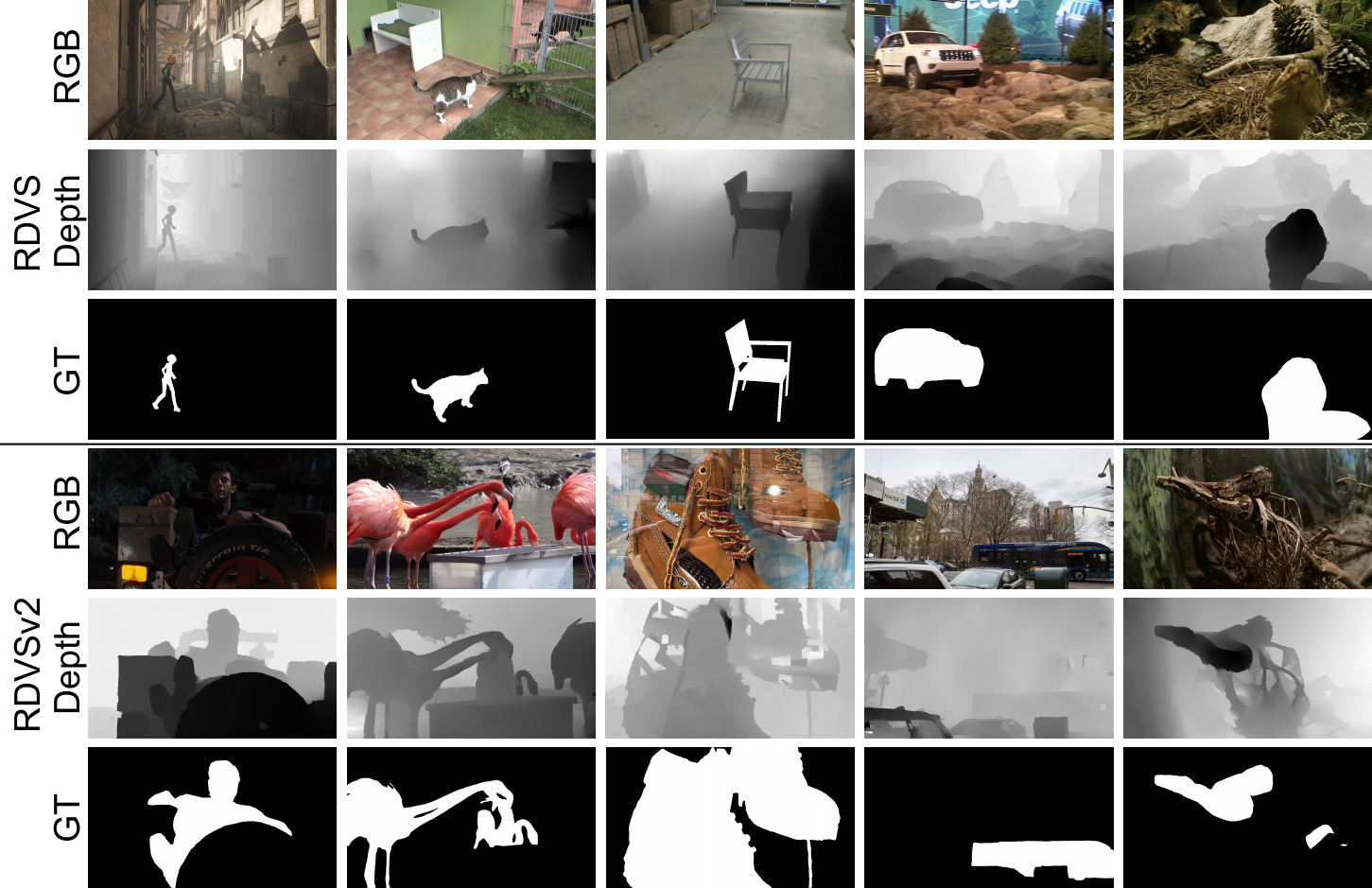}
\vspace{-0.1em}
\caption{
Comparison of sample frames from \ds and RDVS for several representative categories.
}
\label{fig:V1vsV2}
\vspace{-1em}
\end{figure}

\textbf{Attribute Analysis of \ds.} 
To demonstrate that \ds contains diverse and challenging scenarios, we conduct a statistical analysis of 16 commonly used video attributes following the settings of DAVIS~\cite{perazzi2016benchmark} and RDVS. Specifically, we adopt the 14 attributes, including HO (Heterogeneous Object), OCC (Occlusion), OV (Out-of-view), FM (Fast-Motion), MB (Motion Blur), DEF (Deformation), SC (Shape Complexity), SV (Scale-Variation), AC (Appearance Change), BC (Background Clutter), BO (Big Object), SO (Small Object), IN (Indoor), and OUT (Outdoor). We further introduce two additional attributes: SH (Shift) indicating that saliency shifts between objects over time, and MO (Multiple Objects), indicating the presence of multiple salient objects. Detailed definitions of the attributes are provided in the \emph{supplementary material}. As shown in Fig.~\ref{fig:combined} (left), compared to RDVS, \ds exhibits higher counts for most attributes, indicating that, on average, each video is associated with a richer combination of attributes, reflecting higher complexity. 

Furthermore, these attributes are not mutually exclusive. Fig.~\ref{fig:combined} (right) illustrates the pairwise dependencies among these attributes across the entire dataset, revealing rich inter-attribute relationships. This indicates that video sequences in \ds often involve multiple attributes simultaneously rather than a single isolated factor, resulting in more complex and challenging scenarios.

We also select several representative frames from RDVS and \ds for a visual comparison (Fig.~\ref{fig:V1vsV2}). From the GT masks, salient objects in \ds appear more complex in shape and configuration. The RGB frames further show that, even within similar scene categories, \ds tends to contain more intricate scenes. This is also reflected in the depth maps, where the increased scene complexity makes salient regions less visually apparent.

\begin{figure}[t]
\centering
\begin{minipage}[t]{0.99\linewidth}
    \vspace{0pt}
    \centering
    \includegraphics[width=\linewidth]{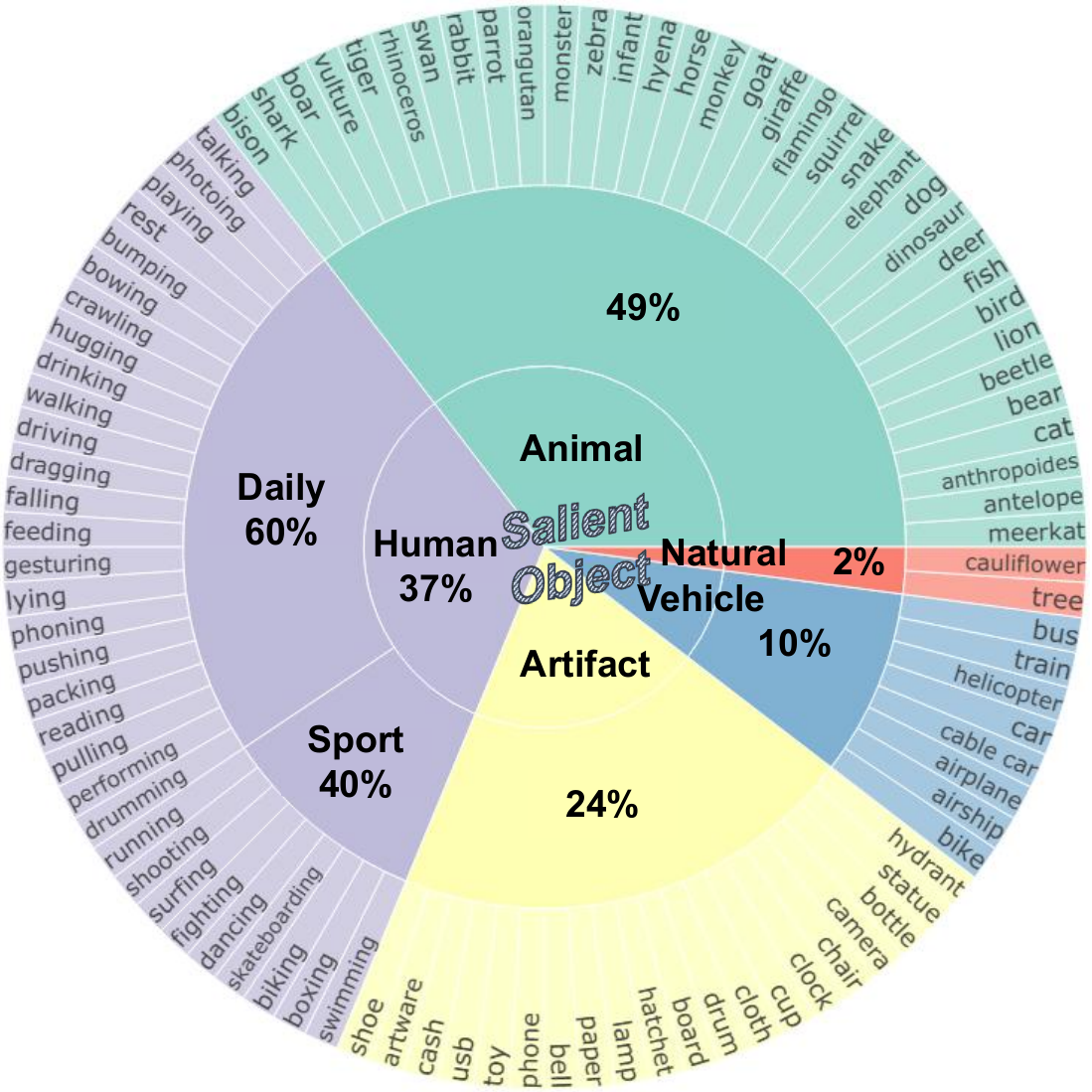}
    \caption{Categories of salient objects}
    \label{fig:salient_object}
\end{minipage}
\par\vspace{0.5em}
\begin{minipage}[t]{0.9\linewidth}
    \vspace{0pt}
    \centering
    \includegraphics[width=\linewidth]{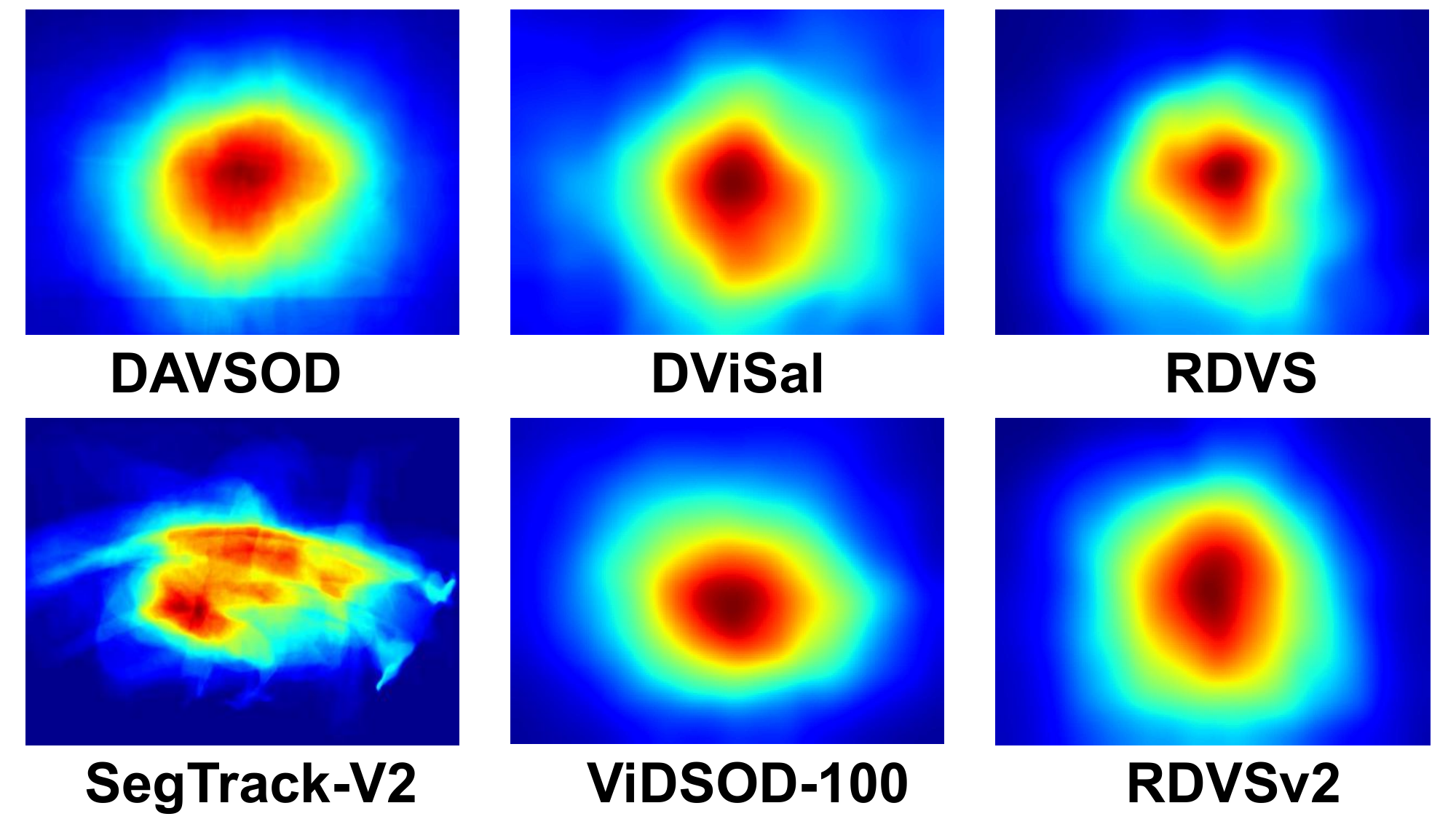}
    \caption{Center bias}
    \label{fig:center_bias}
\end{minipage}
\vspace{-2em}
\end{figure}

\textbf{Object Category Analysis of \ds.} 
Fig.~\ref{fig:salient_object} illustrates the distribution of object categories in \ds. Consistent with~\cite{fan2019shifting, wang2019revisiting, mou2024salient}, we manually annotate each sequence with one or multiple categories (Human, Animal, Vehicle, Artifact, and Natural). Note that these categories are not mutually exclusive; thus, the sum of the percentages exceeds 100\%. Animal is the most prevalent category, appearing in 49\% of video sequences (e.g., lion, bear). Human activities follow at 37\% and can be further divided into two mutually exclusive subcategories: Daily activities account for 60\% (e.g., talking, walking), while Sports account for the remaining 40\% (e.g., running, dancing). Artifacts appear in 24\% of the sequences (e.g., artware, statue), and Vehicles are present in 10\% (e.g., car, bike). Natural objects account for a small proportion, appearing in 2\% of the sequences (e.g., trees). This distribution is broadly consistent with that of RDVS, suggesting that \ds maintains comprehensive coverage of object-level saliency in daily-life scenarios.

Fig.~\ref{fig:center_bias} visualizes center bias through average saliency maps aggregated over all video frames of \ds and several other widely used video based SOD datasets. Because viewers tend to focus on the central region when viewing videos, \ds also exhibits a clear center bias, which is consistent with the patterns observed in datasets such as DAVSOD and RDVS. Moreover, because \ds contains more video frames, its average saliency map appears smoother. This contrasts with SegTrack-V2~\cite{li2013video} and is consistent with the observations reported in RDVS. Further dataset analysis is provided in the \emph{supplementary material}.

\subsection{Dataset Training/Testing Splits}
We consider two evaluation protocols for \ds. The first is a cross-dataset evaluation setting, in which the entire \ds dataset is used for testing.
The second is the official 7:3 training/testing split. Previous RGB-D VSOD datasets~\cite{li2023dvsod, lin2024vidsod} lack a principled training/test split. As a result, the training and test sets may exhibit imbalanced distributions in object categories and video attributes. In contrast, \ds adopts a balanced 7:3 split, with 174 video sequences and 19,499 frames for training, and 75 sequences with 9,578 frames for testing. Object categories and video attributes are also balanced across the two splits under a similar 7:3 ratio. This results in a more balanced division, where both the training and test sets contain a mix of relatively simple and more challenging examples\footnote{Training/test set comparisons are available in the \emph{supplementary material}.}.

\section{Proposed Baseline Model}

\begin{figure*}[t]
  \centering
  \includegraphics[width=0.8\linewidth]{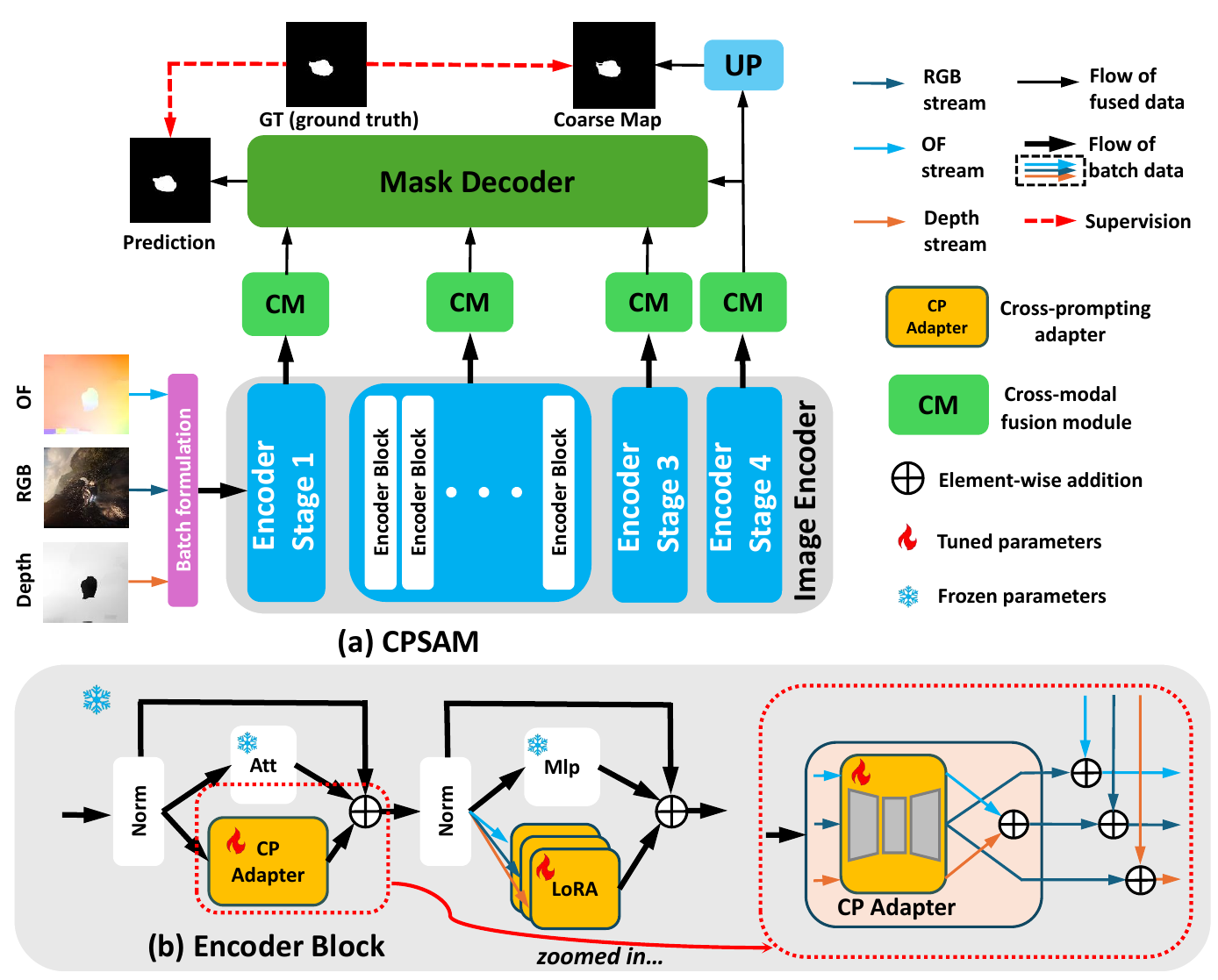}
  \caption{Overview of \model. (a) shows the big picture. (b) shows the details of encoder block in image encoder.}
  \label{fig:model}
\end{figure*}

We adapt SAM2 to RGB-D VSOD and establish a simple baseline based on fine-tuning. Unlike SAM-DAQ, which inputs both RGB and depth into the image encoder and relies on SAM2’s memory mechanism for temporal modeling, our approach adopts a simpler and more general design. Specifically, we only utilize SAM2's image encoder and mask decoder, and feed RGB, depth, and optical flow (from e.g., RAFT, or FlowFormer++~\cite{shi2023flowformer++}) as different modalities to the encoder. As shown in Fig.~\ref{fig:model}, our baseline (\model) integrates a joint-learning cross-modal prompting (CP) adapter and three parallel LoRA~\cite{hu2022lora} modules into each multi-scale block of the image encoder, enabling the model to capture shared and modality-specific information from RGB, depth, and optical flow.

More specifically, as shown in the lower part of Fig.~\ref{fig:model}, since SAM2 is pre-trained only on RGB data, we insert three parallel, modality-specific LoRA modules into the MLP of each encoder block to adapt RGB, depth, and optical flow features separately.
To facilitate multi-modal fusion during fine-tuning, we adopt a \emph{shared} CP adapter inspired by~\cite{liu2025cpal}, which is inserted before the multi-head attention module in each encoder block and residually connected to the attention output. The adapter is implemented as a lightweight bottleneck module composed of three grouped $1\times1$ convolutions. Following previous works, RGB is treated as the primary modality: its output is dispatched to the depth and optical flow branches, while the outputs of the depth and flow branches are merged back into the RGB branch. Before entering the image encoder, the three modalities are formulated into a single batch for joint training.
In addition, the outputs from each encoder stage are fused through a cross-modal fusion module (CM) \cite{fu2020jl} before being passed to the mask decoder for final prediction. 
The fused output from the final stage is upsampled to generate a coarse prediction map, which is supervised to facilitate object localization.

\section{Experiments and Results}
We first evaluate \ds as a challenging benchmark, followed by the performance of the proposed baseline.

\textbf{Datasets.} We consider two experimental settings with different datasets:

To evaluate \ds as a benchmark, \model is trained on three VSOD datasets, namely DAVIS, DAVSOD, and FBMS~\cite{ochs2013segmentation}, which contain 30, 61, and 29 sequences, respectively. For these datasets, monocular depth maps are synthetically generated using DPT~\cite{ranftl2021vision}. In addition, the widely used DUTS~\cite{wang2017learning}, a large-scale static image dataset for SOD, is also included in training. The trained models are first directly evaluated on the full RDVS and \ds. We then fine-tune the existing methods and also \model on the training split of \ds and evaluate them on the \ds test set.

To evaluate the proposed baseline model, we follow the evaluation protocol established in previous work~\cite{suolang2025lightweight} and conduct training and testing separately on three existing RGB-D VSOD datasets: RDVS (57 videos, 4,087 frames), DViSal (237 videos, 7,117 frames), and ViDSOD-100 (100 videos, 9,362 frames).

\textbf{Implementation Details.} \model is built on the large-scale variant of SAM2 (SAM2-L). The input RGB frames, depth maps, and optical flow fields are resized to 1024$\times$1024. For joint multi-modal training, the three modalities are concatenated along the batch dimension, resulting in an effective batch size of 1. During training, the parameters of the SAM2 image encoder are kept frozen. We employ the AdamW optimizer with a learning rate of 2e-5. The model is trained for 30 epochs on a single NVIDIA GeForce RTX 5090 GPU, with a total training time of approximately 24 hours.

\textbf{Evaluation metrics.} We adopt three widely used evaluation metrics, namely S-measure~\cite{fan2017structure}, maximum F-measure~\cite{borji2015salient, achanta2009frequency}, and MAE~\cite{borji2015salient, perazzi2012saliency}. For MAE, the lower score is better, and vice versa for the other metrics. Following the protocol in~\cite{fan2019shifting}, each metric is first averaged over frames within each video, and then averaged over all videos across a dataset.

\subsection{Evaluation of \ds}
To validate the value and complexity of \ds as a benchmark, and to assess the generalizability of algorithms trained on external datasets, we adopt a two-stage evaluation protocol consistent with previous RDVS benchmarking studies.

\begin{table*}[!t]
\centering
\small
\renewcommand{\arraystretch}{1.15}

\caption{\textbf{Comparison on RDVS and \ds.}
``$\uparrow$'' denotes that higher is better.
``$\downarrow$'' denotes that lower is better.
``--'' indicates that the model is trained on the RDVS training set and therefore is not tested on RDVS.
Horizontal lines separate different method categories: RGB-D SOD, VSOD, and RGB-D VSOD.}
\label{tab:rdvs2_comparison}

\vspace{-4pt}

\begin{tabular*}{0.8\textwidth}{
    @{\extracolsep{\fill}}
    l c c c c c c
    @{}
}
\toprule

\multirow{2}{*}{\textbf{Methods}}
& \multicolumn{3}{c}{\textbf{RDVS} \cite{mou2024salient}}
& \multicolumn{3}{c}{\textbf{\ds}}
\\

\cmidrule(lr){2-4}
\cmidrule(lr){5-7}

& $S_{\alpha} \uparrow$
& $F_{\beta}^{\max} \uparrow$
& $M \downarrow$
& $S_{\alpha} \uparrow$
& $F_{\beta}^{\max} \uparrow$
& $M \downarrow$
\\

\midrule

BBSNet \cite{fan2020bbs}
& 0.790 & 0.679 & 0.054
& 0.767 & 0.673 & 0.095
\\

JLDCF \cite{fu2020jl}
& 0.789 & 0.691 & 0.058
& 0.764 & 0.679 & 0.101
\\

SPNet \cite{zhou2021specificity}
& 0.793 & 0.699 & 0.055
& 0.769 & 0.685 & 0.091
\\

RD3D+ \cite{chen20223}
& 0.814 & 0.725 & 0.052
& 0.744 & 0.671 & 0.108
\\

PICRNet \cite{cong2023point}
& 0.787 & 0.693 & 0.059
& 0.757 & 0.672 & 0.097
\\

\midrule

MGAN \cite{li2019motion}
& 0.847 & 0.796 & 0.059
& 0.755 & 0.662 & 0.104
\\

FSNet \cite{ji2021full}
& 0.829 & 0.773 & 0.053
& 0.741 & 0.636 & 0.097
\\

DCFNet \cite{zhang2021dynamic}
& 0.811 & 0.742 & 0.050
& 0.745 & 0.649 & 0.097
\\

\midrule

DCTNet \cite{lu2022depth}
& 0.854 & 0.816 & 0.044
& 0.774 & 0.690 & 0.088
\\

DCTNet+ \cite{mou2024salient}
& 0.871 & 0.836 & 0.042
& 0.789 & 0.707 & 0.087
\\

SAM-DAQ \cite{lin2025sam}
& -- & -- & --
& 0.793 & 0.711 & 0.085
\\

\textbf{CPSAM (Ours)}
& \textbf{0.908}
& \textbf{0.884}
& \textbf{0.029}
& \textbf{0.817}
& \textbf{0.748}
& \textbf{0.077}
\\

\bottomrule
\end{tabular*}

\vspace{-5pt}
\end{table*}

\textbf{Straightforward evaluation on the full \ds dataset.} First, we directly test 11 existing methods~\cite{fan2020bbs, fu2020jl, zhou2021specificity, chen20223, cong2023point, li2019motion, ji2021full, zhang2021dynamic, lu2022depth, mou2024salient, lin2025sam}, each pre-trained on its original training data, on the full \ds and also RDVS datasets. This zero-shot protocol reveals the inherent challenge of \ds and the cross-domain generalization capability of current models.

As shown in Table~\ref{tab:rdvs2_comparison}, the performance of existing models—whether designed for RGB-D SOD, VSOD, or RGB-D VSOD—drops significantly on \ds compared with RDVS, indicating the greater overall complexity of \ds. Specifically, the S-measure scores for VSOD and RGB-D VSOD models drop by approximately 8 percentage points on average.
While VSOD methods outperform RGB-D SOD methods on RDVS, this gap largely disappears on \ds. This suggests that motion cues alone may be less discriminative on \ds. As a result, salient object identification based solely on motion becomes more challenging for VSOD models.
Notably, \model achieves the best overall performance on both benchmarks, demonstrating its strong generalization and robustness.
       
\begin{table*}[!t]
\centering
\small
\renewcommand{\arraystretch}{1.15}

\caption{\textbf{Comparison of original and fine-tuned models on the \ds test set.}}
\label{tab:original_finetuned_comparison}

\vspace{-4pt}

\begin{tabular*}{0.8\textwidth}{
    @{\extracolsep{\fill}}
    l c c c c c c
    @{}
}
\toprule

\multirow{2}{*}{\textbf{Methods}}
& \multicolumn{3}{c}{\textbf{Original model}}
& \multicolumn{3}{c}{\textbf{Fine-tuned model}}
\\

\cmidrule(lr){2-4}
\cmidrule(lr){5-7}

& $S_{\alpha} \uparrow$
& $F_{\beta}^{\max} \uparrow$
& $M \downarrow$
& $S_{\alpha} \uparrow$
& $F_{\beta}^{\max} \uparrow$
& $M \downarrow$
\\

\midrule

BBSNet \cite{fan2020bbs}
& 0.757 & 0.670 & 0.100
& 0.770 & 0.702 & 0.090
\\

JLDCF \cite{fu2020jl}
& 0.760 & 0.679 & 0.104
& 0.769 & 0.701 & 0.103
\\

SPNet \cite{zhou2021specificity}
& 0.772 & 0.695 & 0.090
& 0.791 & 0.734 & 0.082
\\

RD3D+ \cite{chen20223}
& 0.734 & 0.620 & 0.113
& 0.744 & 0.644 & 0.110
\\

PICRNet \cite{cong2023point}
& 0.765 & 0.684 & 0.090
& 0.792 & 0.728 & 0.079
\\

\midrule

MGAN \cite{li2019motion}
& 0.759 & 0.672 & 0.102
& 0.784 & 0.711 & 0.086
\\

FSNet \cite{ji2021full}
& 0.746 & 0.648 & 0.106
& 0.771 & 0.692 & 0.097
\\

DCFNet \cite{zhang2021dynamic}
& 0.738 & 0.650 & 0.098
& 0.769 & 0.695 & 0.086
\\

\midrule

DCTNet \cite{lu2022depth}
& 0.778 & 0.706 & 0.085
& 0.797 & 0.735 & 0.082
\\

DCTNet+ \cite{mou2024salient}
& 0.791 & 0.719 & 0.084
& 0.809 & 0.746 & 0.071
\\

SAM-DAQ \cite{lin2025sam}
& 0.786 & 0.712 & 0.093
& 0.812 & 0.757 & 0.085
\\

\textbf{CPSAM (Ours)}
& \textbf{0.818}
& \textbf{0.759}
& \textbf{0.075}
& \textbf{0.838}
& \textbf{0.791}
& \textbf{0.066}
\\

\bottomrule
\end{tabular*}

\vspace{-5pt}
\end{table*}

\textbf{Evaluation on \ds test set after fine-tuning.} We fine-tune the same 11 methods on the training split of \ds and compare their performance on the test set before and after fine-tuning. This comparison reflects the adaptability of existing methods to \ds.
As shown in Tab.~\ref{tab:original_finetuned_comparison}, methods from all three categories (RGB-D SOD, VSOD, and RGB-D VSOD) show performance gains after fine-tuning, indicating that \ds provides effective supervision for model adaptation. Notably, VSOD and RGB-D VSOD methods exhibit more substantial improvements, with VSOD models gaining approximately 3 percentage points on S-measure, while RGB-D SOD methods show relatively smaller gains. This suggests that the \ds training set is likely to provide more effective adaptation to methods that rely on temporal cues. Furthermore, \model achieves SOTA performance after fine-tuning. More quantitative results on the \ds test set are provided in the \emph{supplementary material}.
Additionally, comparing the original results before fine-tuning on the \ds test set in Tab.~\ref{tab:original_finetuned_comparison} with those on the full \ds dataset in Tab.~\ref{tab:rdvs2_comparison} shows that the performance metrics remain largely consistent. This close agreement further confirms that the training and test splits are balanced and representative.

\begin{figure}[t]
\centering
\includegraphics[width=0.9\linewidth]{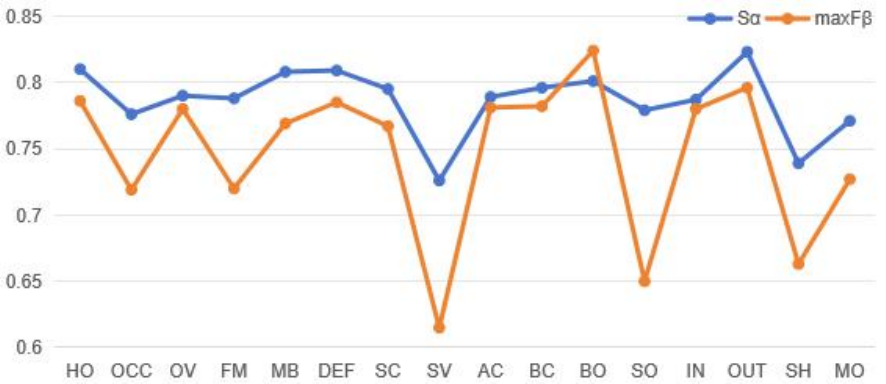}
\vspace{-0.5em}
\caption{
Attribute impact on the difficulty of video sequences. The two metrics are the average of four models.
}
\label{fig:sf}
\vspace{-2em}
\end{figure}

Furthermore, we analyze the performance of four RGB-D VSOD models fine-tuned on the \ds training set at the per-sequence level on the test set. This analysis examines how the attributes introduced in Section~\ref{sec:DSC} affect dataset difficulty. As shown in Fig.~\ref{fig:sf}, models achieve lower S-measure and max F-measure scores on sequences with attributes such as OCC, FM, SV, SO, SH, and MO. Notably, the performance drop is particularly pronounced for sequences with SV, indicating that scale variation is especially challenging for current models. This suggests that incorporating more sequences with such challenging attributes may further increase dataset difficulty in future dataset construction.

\subsection{Extended Evaluation of \model} 
As in the previous subsection, \model is also evaluated in the benchmark study on \ds. In this part, we further benchmark \model against SOTA methods on other RGB-D VSOD datasets.

\begin{table*}[!t]
\centering
\small
\renewcommand{\arraystretch}{1.15}

\caption{\textbf{Comparison with SOTA methods on three existing RGB-D VSOD datasets.}}
\label{tab:comparison_sota}

\vspace{-4pt}

\begin{tabular*}{0.9\textwidth}{
    @{\extracolsep{\fill}}
    l c c c c c c c c c
    @{}
}
\toprule

\multirow{2}{*}{\textbf{Methods}}
& \multicolumn{3}{c}{\textbf{RDVS} \cite{mou2024salient}}
& \multicolumn{3}{c}{\textbf{DViSal} \cite{li2023dvsod}}
& \multicolumn{3}{c}{\textbf{ViDSOD-100} \cite{lin2024vidsod}}
\\

\cmidrule(lr){2-4}
\cmidrule(lr){5-7}
\cmidrule(lr){8-10}

& $S_{\alpha}\uparrow$
& $F_{\beta}^{\max}\uparrow$
& $M\downarrow$
& $S_{\alpha}\uparrow$
& $F_{\beta}^{\max}\uparrow$
& $M\downarrow$
& $S_{\alpha}\uparrow$
& $F_{\beta}^{\max}\uparrow$
& $M\downarrow$
\\

\midrule

DVSOD \cite{fan2019shifting}
& 0.602 & 0.381 & 0.108
& 0.729 & 0.610 & 0.113
& 0.771 & 0.686 & 0.069
\\

ATFNet \cite{lin2024vidsod}
& 0.750 & 0.613 & 0.051
& 0.737 & 0.677 & 0.087
& 0.809 & 0.787 & 0.045
\\

DCTNet+ \cite{mou2024salient}
& 0.868 & 0.813 & 0.034
& 0.794 & 0.770 & 0.073
& 0.880 & 0.832 & 0.040
\\

MFENet \cite{suolang2025lightweight}
& 0.794 & 0.700 & 0.049
& 0.760 & 0.717 & 0.080
& 0.831 & 0.763 & 0.040
\\

Samba \cite{he2025samba}
& 0.879 & 0.826 & 0.029
& 0.855 & 0.841 & 0.042
& 0.924 & 0.896 & \textbf{0.014}
\\

SAM-DAQ \cite{lin2025sam}
& 0.877 & 0.821 & 0.035
& 0.855 & 0.850 & 0.042
& 0.901 & 0.869 & 0.026
\\

\textbf{CPSAM (Ours)}
& \textbf{0.899}
& \textbf{0.856}
& \textbf{0.028}
& \textbf{0.879}
& \textbf{0.879}
& \textbf{0.030}
& \textbf{0.926}
& \textbf{0.897}
& \textbf{0.014}
\\

\bottomrule
\end{tabular*}

\vspace{-5pt}
\end{table*}

\textbf{Comparison with SOTA Methods.} We compare \model with 6 SOTA RGB-D VSOD approaches~\cite{fan2019shifting, lin2024vidsod, mou2024salient, suolang2025lightweight, he2025samba, lin2025sam}. The quantitative results on three established RGB-D VSOD datasets are presented in Table~\ref{tab:comparison_sota}. \model achieves the best overall performance among the compared methods. Moreover, compared with SAM‑DAQ—the previous best‑performing method that also uses SAM2 as backbone—\model achieves average improvements of 2.4\% in S‑measure, 3.1\% in max F‑measure, and 0.01 in MAE. It also outperforms Samba, a Mamba-based method, across all three datasets. These quantitative comparisons demonstrate the effectiveness and competitiveness of the proposed baseline. 

\begin{table}[t]
\centering
\scriptsize
\caption{\textbf{Ablation study on RDVS \cite{mou2024salient}.}}
\vspace{-1em}
\label{tab:ablation}
\setlength{\tabcolsep}{8.0pt}
\renewcommand{\arraystretch}{1.08}
\begin{tabular}{l*{5}{c}}
\toprule
\textbf{Metric} & \textbf{A1} & \textbf{A2} & \textbf{A3} & \textbf{A4} &  \textbf{CPSAM} \\
\midrule
$S_{\alpha} \uparrow$   & 0.846 & 0.862 & 0.878 &  0.886 & \textbf{0.899} \\
$F_{\beta}^{\max} \uparrow$   & 0.772 & 0.803 & 0.832 & 0.851 & \textbf{0.856} \\
$M \downarrow$   & 0.037 & 0.034 & 0.035 & 0.031 & \textbf{0.028} \\
\bottomrule
\end{tabular}
\vspace{-1.5em}
\end{table}

\textbf{Ablation Study.} We evaluate the CP adapter and parallel LoRA in \model on RDVS. Four variants are considered: A1 (both removed), A2 (without parallel LoRA), A3 (without the CP adapter), and A4 (using a single LoRA instead of the three parallel LoRA module). As shown in Table~\ref{tab:ablation}, compared with A1, A2 yields limited gains, while A3 brings clear improvements, indicating that parallel LoRA helps capture modality-specific differences and improves SAM2’s adaptation to depth and optical flow. A4 also performs worse than the full model, further confirming the effectiveness of parallel LoRA. The best performance is achieved when the CP adapter and parallel LoRA are used together, suggesting that the two components are complementary. More ablation results are provided in the \emph{supplementary material}.

\section{Conclusion} 
In this work, we address two important challenges in RGB-D video salient object detection: the lack of a large-scale, high-quality benchmark and the need for an effective baseline adapted from vision foundation models to such a multi-modal setting. We introduce \ds, a large-scale benchmark featuring stereo-derived depth and eye-tracking-guided masks, with improved scale and scene complexity over existing RGB-D VSOD datasets. We also present a parameter-efficient adaptation strategy for SAM2 that combines modality-specific LoRA with a shared cross-modal prompting adapter, establishing an effective baseline for RGB-D VSOD. Extensive experiments show the challenging nature of \ds and demonstrate the strong performance of \model compared with existing SOTA methods. We hope that \ds and the provided baseline will facilitate future research in the community and could be extended to other related multi-modal video understanding tasks.

\bibliographystyle{unsrt}
\balance
\bibliography{custom}

\end{document}